\documentclass[11pt]{article}
\usepackage{coling2018}
\usepackage{times}
\usepackage{url}
\usepackage{latexsym}
\usepackage{amsmath,amssymb}
\usepackage{graphicx}
\usepackage{makecell}

\DeclareMathOperator*{\argmax}{arg\,max}

\usepackage{xcolor}

\newcommand{\nb}[1]{}

\title{SeVeN: Augmenting Word Embeddings\\ with Unsupervised Relation Vectors}

\author{Luis Espinosa-Anke and Steven Schockaert \\
  School of Computer Science and Informatics, 
  Cardiff University, UK \\
  {\tt \{Espinosa-AnkeL,SchockaertS1\}@cardiff.ac.uk}}

\date{}

\begin{document}
\maketitle
\begin{abstract}
%\todo{}
We present SeVeN (Semantic Vector Networks), a hybrid resource that encodes relationships between words in the form of a graph. Different from traditional semantic networks, these relations are represented as vectors in a continuous vector space. We propose a simple pipeline for learning such relation vectors, which is based on word vector averaging in combination with an \textit{ad hoc} autoencoder.
%With this resource we aim at bridging the gap between the hard task to model discrete relations between words and entities in resources like WordNet or DBpedia, and the semantic and syntactic information that is naturally preserverd in word embeddings. 
We show that by explicitly encoding relational information in a dedicated vector space we can capture aspects of word meaning that are complementary to what is captured by word embeddings. For example, by examining clusters of relation vectors, we observe that relational similarities can be identified at a more abstract level than with traditional word vector differences. Finally, we test the effectiveness of semantic vector networks in two tasks: measuring word similarity and neural text categorization. SeVeN is available at \url{bitbucket.org/luisespinosa/seven}\blfootnote{
    %
    % for review submission
    %
    %\hspace{-0.65cm}  % space normally used by the marker
    %Place licence statement here for the camera-ready version. See
    %Section~\ref{licence} of the instructions for preparing a
    %manuscript.
    %
    % % final paper: en-uk version 
    
     \hspace{-0.65cm}  % space normally used by the marker
     This work is licensed under a Creative Commons 
     Attribution 4.0 International Licence.
     Licence details:
     \url{http://creativecommons.org/licenses/by/4.0/}.
     
    % % final paper: en-us version 
    %
    % \hspace{-0.65cm}  % space normally used by the marker
    % This work is licensed under a Creative Commons 
    % Attribution 4.0 International License.
    % License details:
    % \url{http://creativecommons.org/licenses/by/4.0/}.
}.
\end{abstract}

\section{Introduction}
Word embedding models such as Skip-gram \cite{DBLP:conf/nips/MikolovSCCD13} and GloVe \cite{glove2014} use fixed-dimensional vectors to represent the meaning of words. These word vectors essentially capture a kind of similarity structure, which has proven to be useful in a wide range of Natural Language Processing (NLP) tasks. Today, one of the major applications of word embeddings is their interaction with neural network architectures, enabling a kind of generalization beyond those words that were only observed during training. For example, if a classification model has learned that news stories containing words such as `cinema', `restaurant' and `zoo' tend to be categorized as `entertainment', it may predict this latter label also for stories about theme parks due to the shared semantic properties encoded in word vectors. Word embeddings thus endow neural models with some form of world knowledge, without which they would be far less effective. This has prompted a prolific line of research focused on improving word embeddings not only with algorithmic sophistication, but also via explicit incorporation of external knowledge sources such as WordNet \cite{DBLP:conf/naacl/FaruquiDJDHS15}, BabelNet \cite{camacho2016nasari} or ConceptNet \cite{Speeretal2016}.

Regardless of how word vectors are learned, however, the use of fixed-dimensional representations constrains the kind of knowledge they can encode. Essentially, we can think of a word vector as a compact encoding of the salient attributes of the given word. For instance, the vector representation of \textit{lion} might implicitly encode that this word is a noun, and that lions have attributes such as `dangerous', `predator' and `carnivorous'. Beyond these properties, word embeddings can also encode \emph{relational knowledge}. For instance, the embedding might tell us that the words `lion' and `zebra' are semantically related, which together with the attributional knowledge that lions are predators and zebras are prey may allow us to plausibly infer that `lions eat zebras'. However, the way in which relational knowledge can be encoded in word embeddings is inherently limited. One issue is that only relationships which are sufficiently salient can affect the vector representations of their arguments; e.g.\ the fact that Trump has visited France is perhaps not important enough to be encoded in the embeddings of the words `Trump' and `France' (i.e.\ there may be insufficient corpus-based evidence on this fact). Note that this is not a matter of how the embedding is learned; forcing the vector representations to encode this fact would distort the similarity structure of the embedding. From a formal point of view, there are also severe limitations to what can be encoded \cite{gutierrez2018knowledge}. As a simple example, methods based on vector translations cannot model symmetric relations, and they are limited in the kind of many-to-many relations that can be encoded \cite{DBLP:conf/aaai/LinLSLZ15}. 

Like word embeddings, semantic networks such as WordNet \cite{Miller1995}, BabelNet \cite{navigli2012babelnet} or ConceptNet \cite{Speeretal2016} also encode lexical and world knowledge. They use a graph representation in which nodes correspond to words, phrases, entities or word senses. Edge labels are typically chosen from a small set of discrete and well-defined lexical and ontological relationships. Compared to word embeddings, the knowledge captured in such resources is more explicit, and more focused on relational knowledge (although attributional knowledge can be encoded as well, e.g., by using edge labels modeling the \texttt{has\_property} relation). The use of discrete labels for encoding relation types, however, makes such representations too coarse-grained for many applications (e.g., a large proportion of the edges in ConceptNet are labelled with the generic `related to' relationship). It also means that subjective knowledge cannot be modeled in an adequate way (e.g., forcing us to make a hard choice between which animals are considered to have the property `dangerous' and which ones do not). %Finally, the use of discrete edge labels also makes such networks difficult to learn from data. %{\color{blue} worth mention here that acl pilehvar et al. (2017) find that injecting wordnet information to neural classifiers is too *fine-grained* and leads to sparsity?}

In this paper, we propose a hybrid representation, which we call SeVeN (Semantic Vector Networks). Similar to semantic networks, we use a graph based representation in which nodes are associated with words. In contrast to semantic networks, however, these edges are labelled with a vector, meaning that relation types are modeled in a continuous space. 

To obtain a suitable \emph{relation vector} for two given words $a$ and $b$, we start by averaging the vector representations (from a pre-trained word embedding) of the words that appear in sentences that mention both $a$ and $b$. The resulting vectors have two main disadvantages, however. First, they are high-dimensional, as they are constructed as the concatenation of several averaged word vectors.
%we construct separate vectors for the words appearing before $a$, in between $a$ and $b$, and after $b$, and we separately construct such vectors from sentences where $a$ appears befor $b$ and from sentences where $b$ appears before $a$. 
Second, the relation vectors are influenced by words that describe the relationship between $a$ and $b$, but also by words that rather relate to the individual words $a$ or $b$ (as well as some non-informative words). Intuitively we want to obtain a vector representation which only reflects the words that relate to the relationship. For example, the relation vector for (paris,france) should ideally be the same as the vector for (rome,italy), but this will not be the case for the averaged word vectors, as the former relation vector  will also reflect the fact that these words represent places and that they relate to France. To address both issues, we introduce an autoencoder architecture in which the input to the decoder comprises both the encoded relation vector and the word vectors for $a$ and $b$. By explicitly feeding the word vectors for $a$ and $b$ into the decoder, we effectively encourage the encoder to focus on words that describe the relationship between $a$ and $b$.

Once the semantic vector network has been learned, it can be used in various ways. For instance, the relation vectors could be used for measuring relational similarity \cite{DBLP:conf/semeval/JurgensMTH12}, for identifying words that have a specific lexical relationship such as hypernyms \cite{Vylomova2016}, or complementing open information extraction systems \cite{dellibovietal:2015}. In this paper, however, we will assess the potential of SeVeN in terms of two tasks, namely using it for (1) unsupervised semantic similarity modeling, and for (2) enriching word vectors as the input to neural network architectures. The overarching idea in the latter case is that, instead of simply representing each word by its vector representation, the representation for each word position will be composed of (i) the vector representation of the word, (ii) the vector representations of the adjacent words in the semantic vector network, and (iii) the corresponding relation vectors (i.e.\ the edge labels). %Our main hypothesis is that augmenting pre-trained word vectors with vectors from a pre-trained semantic vector network, while keeping the neural network architecture of choice fixed, can lead to considerable gains. % thanks to the additional world knowledge that is implicitly captured in the input representations.

%******************************************************************************************************************************
\section{Related Work}

Related work broadly falls in two categories: methods which aim to improve word embeddings using relational knowledge, and methods which aim to learn relation vectors. To the best of our knowledge, there is no previous work which uses relation vectors with the aim of enriching word embeddings.\smallskip

\noindent \textbf{Improving Word Embeddings.} One of the most notable features of word embedding models, such as Skip-gram \cite{mikolov2013linguistic} and GloVe \cite{glove2014}, is the fact that various syntactic and semantic relationships approximately correspond to vector translations. One limitation of vector translations is that they are not well-suited for modeling transitive relations, which is problematic among others for the is-a relationship. To this end, a number of alternative vector space representations have been proposed, which are specifically aimed at modeling taxonomic relationships \cite{DBLP:journals/corr/VendrovKFU15,yu2015learning,nickel2017poincare}. Note that while such alternative embedding spaces can solve some of the limitations of standard embeddings w.r.t.\ modeling taxonomic relationships, there are many other types of relations that cannot be faithfully modeled in these representations. Moreover, these alternative embeddings are not necessarily well-suited for modeling word similarity. More generally, various authors have explored the idea of adapting word embeddings to fit the needs of specific tasks, e.g.\ aiming to make embeddings better suited for capturing antonyms \cite{ono2015word}, hypernyms \cite{DBLP:journals/corr/abs-1710-06371} or sentiment \cite{tang2014learning}.

As mentioned in the introduction, the use of semantic networks for improving word embeddings, based on the idea that words which are similar in the semantic network should have a similar embedding, has been explored by various authors \cite{DBLP:conf/naacl/FaruquiDJDHS15,camacho2016nasari,Speeretal2016}. Another possibility is to use a semantic network to decompose word embeddings into sense embeddings by imposing the constraint that the word vector is a convex combination of the corresponding sense vectors, as well as  forcing similarity of the sense vector with the vector representations of its neighbors in the semantic network \cite{DBLP:conf/naacl/JohanssonP15}. Finally, let us refer to work that learns additional embeddings that coexist in the same space as lexemes, e.g., WordNet synsets \cite{rothe2015autoextend} or BabelNet synsets \cite{mancini2017embedding}. \smallskip

\noindent \textbf{Relation Vectors}
The idea of learning a relation vector for two words $a$ and $b$, based on the words that appear in their context, goes back at least to the Latent Relational Analysis (LRA) method from \cite{Turney:2005:MSS:1642293.1642475}. In that work, a matrix is constructed with one row for each considered word pair, where columns correspond to lexical patterns that have been extracted from sentences containing these words. The relation vectors are then obtained by applying Singular Value Decomposition (SVD) on that matrix. Along similar lines, in \cite{DBLP:conf/naacl/RiedelYMM13} relation vectors are learned by factorizing a matrix whose rows correspond to entity pairs and whose columns correspond to properties (in this case comprising both lexical patterns from a corpus and triples from a knowledge graph). More recently, several methods have been proposed that learn a vector describing the relationship between two words by averaging the embeddings of the words that appear in between them in a given corpus \cite{DBLP:conf/emnlp/WestonBYU13,DBLP:conf/conll/HashimotoSMT15,DBLP:conf/ranlp/FanCHG15}, or by learning a vector representation from PMI-like statistics on how strongly different words are associated with the considered word pair \cite{DBLP:journals/corr/abs-1711-05294}. Beyond these unsupervised methods, a wide variety of supervised neural network based architectures have been proposed for learning relation vectors that are predictive of a given relation type \cite{zeng2014relation,DBLP:conf/acl/SantosXZ15,xu2015classifying}. 

%{\color{blue} what is missing in the above methods? what sets us apart?}

% word vector averaging & arxiv paper Shoaib

%******************************************************************************************************************************
\section{Constructing Semantic Vector Networks}
Our aim is to construct a graph whose nodes correspond to words, whose edges indicate which words are related, and whose edge labels are vectors that encode the specific relationship between the corresponding words. We will refer to this representation as a semantic vector network. In this section, we describe our methodology for constructing such semantic vector networks. First, in Section \ref{secStructure}, we provide details about the source corpus and explain how the structure of the network is chosen. In Section \ref{secLearningVectors} we then discuss how suitable relation vectors can be constructed. 

%--------------------
\subsection{Defining the Network Structure}\label{secStructure}
%\todo{Discuss how the corpus is preprocessed and tokenized, how multi-word expressions are identified, and which stopwords are removed}.
Our source corpus is a dump of the English Wikipedia from January 2018. We opted to keep preprocessing at a minimum to ensure that any emergent linguistic or relational regularity is captured during the network construction stages. Specifically, we applied sentence segmentation and word tokenization using \textit{nltk}\footnote{\url{nltk.org}}. We also single-tokenized multiword expressions based on several lexicons \cite{schneider2014discriminative}, and finally removed stopwords using the CoreNLP list\footnote{\url{github.com/stanfordnlp/CoreNLP/blob/master/data/edu/stanford/nlp/patterns/surface/stopwords.txt}}\nb{Did we include stopwords for learning the relation vectors? If not then we would miss important ones such as "such", "as", "at", "of", or are these not included?}.
After the above steps, we selected the $10^5$ most frequent words as our vocabulary. To determine which words should be connected with an edge, we rely on Pointwise Mutual Information (PMI), which measures the strength of association between two random variables. It is commonly used in NLP as a method for identifying related words, e.g.\ in factorization based methods for learning word embeddings \cite{turney2010frequency}. Specifically, we express the strength of association between words $w_i$ and $w_j$ as follows:
$$
\textit{pmi}(w_i,w_j) = \log\left(\frac{x_{ij} x_* }{x_{i}x_j} \right)
$$
In our case, $x_{ij}$ is the number of times word $w_i$ appears near word $w_j$, weighted by the nearness of their co-occurrences, and $x_i=\sum_j$, $x_{ij}=\sum_j x_{ji}$ and $x_*=\sum_i\sum_j x_{ij}$. Specifically, let $I_{w_i}$ be the word positions in the corpus at which $w_i$ occurs, then we define:
\begin{align*}
x_{ij} = \sum_{p \in I_{w_i}}\sum_{q \in I_{w_j}} n(p,q)   
\end{align*}
where $n(p,q)=0$ if the word positions $p$ and $q$ belong to a different sentence, or if $|p-q|>10$, i.e.\ if there are at least 10 words in between them. Otherwise we define $n(p,q) = \frac{1}{|p-q|}$.

\begin{table}[!h]
\centering
\resizebox{\textwidth}{!}{
\begin{tabular}{cc|cc|cc|cc}
\multicolumn{2}{c|}{\texttt{\textbf{sorrow}}}      & \multicolumn{2}{c|}{\texttt{\textbf{tournament}}}       & \multicolumn{2}{c|}{\texttt{\textbf{videogame}}}    & \multicolumn{2}{c}{\texttt{\textbf{riverbank}}} \\ \hline \hline
\texttt{ppmi}        & \texttt{w2v}            & \texttt{ppmi}               & \texttt{w2v}       & \texttt{ppmi}        & \texttt{w2v}              & \texttt{ppmi}          & \texttt{w2v}      \\ \hline
contrition  & sadness           & scotties           & tourney      & lego        & videogames          & danube        & riverbanks  \\
lamentation & anguish           & double-elimination & tournaments  & consoles    & videogaming         & erosion       & river       \\
woe         & grief             & single-elimination & Tournament   & villains    & next\_gen\_consoles & laboratories  & creek       \\
savior      & profound\_sorrow  & pre-olympic        & tournment    & arcade      & Videogame           & opposite      & riverbed    \\
everlasting & deepest\_sorrow   & 4-day              & tourament    & sega        & gamers              & vegetation    & riverside   \\
anguish     & heartfelt\_sorrow & eight-team         & tourneys     & ea          & MMOG                & tales         & lake        \\
grief       & profound\_sadness & winnings           & touranment   & playstation & PS2                 & washed        & shoreline   \\
%sadness     & despair           & round-robin        & tournement   & gran        & Xbox                & downstream    & creek\_bed  \\
%castlevania & sorrowful         & interzonal         & championship & console     & PlayStation         & steep         & mud\_flats  \\
%joy         & saddness          & tri-nations        & quarterfinal & developer   & Tony\_Hawk\_RIDE    & riverside     & canal      
\end{tabular}
}
\caption{Examples of the highest scoring (i.e.\ most strongly associated by \texttt{pmi}) words,  as well as their nearest neighbors in the pretrained word2vec (\texttt{w2v}) Google news vector space.}
\label{tab:bestedges}
\end{table}

To choose the edges of the semantic vector network, we only consider word pairs which co-occur at least 10 times in the corpus. Among such pairs, for each word $w_i$, we first select the 10 words $w_j$ whose score $\textit{pmi}(w_i,w_j)$ is highest. This resulted in a total of about 900\,000 pairs. Then, we added the overall highest scoring pairs $(w_i,w_j)$ which had not yet been selected, until we ended up with a total of approximately $10^6$ edges involving the initial vocabulary of $10^5$ words. In the following we will write $N_w$ for the neighbors of $w$, i.e.\ the set of words $n$ such that $\{w,n\}$ was selected as an edge. %Note that this set of neighbors contains words that were selected because they are among the most closely related words to $w$, as well as words for which $w$ was selected as one of the most closely related words.

Note that by capturing pairs of words strongly connected by PMI, we encode a different type of relatedness than proximity in word embeddings. 
%We can think of the PMI-derived neighborhood as encoding a more paradigmatic relation, i.e., beyond the sequential nature of syntagmatic relations captured in the skip-gram model, for instance, and also preventing clustering together similar spellings due to them sharing similar contexts. 
To illustrate this, in Table \ref{tab:bestedges} we compare the most closely related words in our PMI graph, for some selected target words, with their nearest neighbors in the \textit{word2vec} Google News word embedding space\footnote{\url{https://code.google.com/archive/p/word2vec/}} (measured by cosine similarity). 
While the word2vec neighbors mostly consist of near-synonyms and other syntagmatic relationships, the chosen PMI pairs include a wide variety of topically related linguistic items.
%We can clearly see a stronger tendency of PMI association to group topically related linguistic items. %, intuitively providing a solid ground for which relational knowledge would be useful to encode. 
A semantic network based on such PMI pairs should thus capture information which is complementary to what is captured in word embeddings.
% and which may clearly be of benefit to an NLP system.
%For example, it may be important to know that tournaments can be played single or double-elimination brackets, or that videogames are related to villains.

%For illustrative purposes, in Table \ref{tab:bestedges} we show five sample words\nb{Can we replace the Cardiff example with another one? It sort of gives away something about who wrote the paper. Also, since we have enough space left, we may as well show the top 10 neighbors} and their five most related edges, as per the above procedure.

%\todo{If there is time and space left, perhaps add table with some examples of words, and their top edges?}. 

%--------------------
\subsection{Learning relation vectors}\label{secLearningVectors}

Our general strategy for learning relation vectors is based on averaging word vectors. Specifically, for each sentence $s$ in which $w_i$ occurs before $w_j$ (within a distance of at most 10), we construct three vectors, based on the words $a_1,...,a_k$ which appear before $w_i$, the words $b_1,...,b_l$ which appear in between $w_i$ and $w_j$ and the words $c_1,...,c_q$ which appear after $w_j$:%\nb{Altough we take a maximum of 10 words into account before $w_i$ and after $w_j$?}: 
\begin{align*}
\textit{pre}^s_{w_iw_j} &= \frac{1}{k} \sum_{r=1}^k \mathbf{v}_{a_r} &
\textit{mid}^s_{w_iw_j} &= \frac{1}{l} \sum_{r=1}^l \mathbf{v}_{b_r} &
\textit{post}^s_{w_iw_j} &= \frac{1}{q} \sum_{r=1}^q \mathbf{v}_{c_r}
\end{align*}
where we write $\mathbf{v}_w$ for the vector representation of the word $w$. These vectors are then averaged over all sentences $S_{ij}$ where $w_i$ occurs before $w_j$:
\begin{align*}
\textit{pre}_{w_iw_j} &= \frac{1}{|S_{ij}|} \sum_{s\in S_{ij}} \textit{pre}^s_{w_iw_j} &
\textit{mid}_{w_iw_j} &= \frac{1}{|S_{ij}|} \sum_{s\in S_{ij}} \textit{mid}^s_{w_iw_j} &
\textit{post}_{w_iw_j} &= \frac{1}{|S_{ij}|} \sum_{s\in S_{ij}} \textit{post}^s_{w_iw_j}
\end{align*}
Since we can similarly obtain such vectors from sentences where $w_j$ appears before $w_i$, we end up with a relation vector whose dimensionality is six times higher than the dimensionality of the word embedding, which would be impractical in the kind of applications we envisage (see Section \ref{secEvaluation}). Another problem with these vectors is that they do not only reflect the relationship between $w_i$ and $w_j$, but also the words $w_i$ and $w_j$ themselves. For instance, suppose we want to model the relationship between the words `movie' and `popcorn'. A sentence mentioning these two words could be:
\begin{quote}
\textit{Buttered popcorn is commonly eaten at movie theatres.}
\end{quote}
The most relevant words for describing the relationship are `eaten at'. In contrast, however, `buttered' is mostly related to the word `popcorn' itself rather than describing its relationship with `movie'. Similarly, `theatres' is related to `movie', but not relevant for characterizing the relationship. 

To solve both issues, we propose to use an autoencoder architecture, in which the decoder has access to the word vectors $\mathbf{v}_{w_i}$ and $\mathbf{v}_{w_j}$, in addition to the encoded version of the relation vector. Let us write $\mathbf{z}_{w_iw_j}$ for the concatenation of $\textit{pre}_{w_iw_j}$, $\textit{mid}_{w_iw_j}$, $\textit{post}_{w_iw_j}$, $\textit{pre}_{w_jw_i}$, $\textit{mid}_{w_jw_i}$, $\textit{post}_{w_jw_i}$. Then the encoder is given by:
$$
\mathbf{r}_{w_iw_j} = A\, \mathbf{z}_{w_iw_j} + \mathbf{b}
$$
where $A \in \mathbb{R}^m \times \mathbb{R}^{6d}$ and $\mathbf{b}\in \mathbb{R}^m$, where $d$ is the dimensionality of the word vectors, and $m$ is the dimensionality of the encoded relation vectors $m$. In our experiments we experiment with different values for $m$. Empirically, we find that as the dimensionality of the compressed representations becomes smaller, the importance of word semantics gradually fades away in favor of their corresponding relational properties. Then, the decoder is then defined as:
$$
\mathbf{z}_{w_iw_j}^* = B (\mathbf{v}_{w_i} \oplus \mathbf{r}_{w_iw_j} \oplus \mathbf{v}_{w_j}) + \mathbf{c}
$$
where $\oplus$ denotes vector concatenation, $B \in \mathbb{R}^{6d} \times \mathbb{R}^{m+2d}$ and $\mathbf{c}\in \mathbb{R}^{6d}$.
To train the autoencoder, we use the following L2-regularized reconstruction loss:
$$
\mathcal{L} = \|\mathbf{z}_{w_iw_j} - \mathbf{z}_{w_iw_j}^* \|_2^2 + \lambda \|\mathbf{r}_{w_1n_1}\|_2^2
$$ 
with $\lambda>0$ a regularization parameter. This loss function balances two objectives: minimizing the reconstruction error and keeping the L2 norms of the encoded relation vectors as small as possible. Because of this latter part, we can think of the norm of the relation vectors $\mathbf{r}_{w_iw_j}$ as a measure of how strongly the words $w_i$ and $w_j$ are related. In particular, if sentences mentioning $w_i$ and $w_j$ contain few or no words that describe their relationship, we might expect $\mathbf{r}_{w_iw_j}$ to be close to the 0 vector.\footnote{We also conducted experiments without the regularization term, with slightly worse results across all evaluations.}

%Note that while the structure of the graph is undirected, in the sense that $w_i$ is a neighbor of $w_j$ iff $w_j$ is a neighbor of $w_i$, the edge labels are directed, in the sense that $\mathbf{r}_{w_iw_j}\neq \mathbf{r}_{w_jw_i}$. This is indeed desirable since we want to be able to model asymmetric relationships.

% The figure should reflect the text notation when describing the autoencoder (e.g., we may want each partial context to be denoted as c_lx, c_cx, c_rx, c_ly, c_cy and c_ry, where x and y are source and target words. Any feedback on the figure can go here.
%\begin{center}
%\begin{figure}[!t]
%\centering
%\includegraphics[width=175pt]{encodernewSteven}
%\caption{Our proposed autoencoder architecture. A standard autoencoder would attempt to reconstruct the input relation vector given a compressed representation. In our architecture, the decoder is explicitly fed the source and target words of the given relation to ensure that the compressed representation is focused on the relationship between the two words.}
%\label{fig:autoencoder}
%\end{figure}
%\end{center}

%******************************************************************************************************************************
\section{Evaluation}\label{secEvaluation}

We propose to evaluate our semantic vector networks from three different standpoints. First, we provide a qualitative evaluation by exploring relation network spaces (both compressed and uncompressed) and discussing meaningful properties. Second, we perform experiments in word similarity where we compare against the standard approach of measuring the similarity of two words by means of the cosine distance between their corresponding word vectors. This evaluation serves as an illustration of how semantic vector networks could be used in an unsupervised application setting. Third, as a prototypical example of a supervised application setting, we analyze the impact of leveraging the enriched representation these networks provide in neural text classification, in particular topic categorization and sentiment analysis. In all experiments, the pretrained embeddings we use (both for baselines and for constructing the relation networks) are the \textit{word2vec} Google News embeddings \cite{mikolov2013linguistic}.

\subsection{Qualitative Evaluation}

One of the strongest selling points of word embeddings is that they enable inference of relational properties, which can be obtained by simple vector arithmetic such as summation and subtraction \cite{levy2015improving}. The basic idea is that the relationship between two words $w_i$ and $w_j$ is characterized by the vector difference $\mathbf{v}_{w_i} - \mathbf{v}_{w_j}$. Such vector differences, however, encode relations in a noisy way. For instance, while the differences $\mathbf{v}_{\textit{rome}} - \mathbf{v}_{\textit{italy}}$, $\mathbf{v}_{\textit{paris}} - \mathbf{v}_{\textit{france}}$ and $\mathbf{v}_{\textit{dublin}} - \mathbf{v}_{\textit{ireland}}$ are all rather similar, there are in fact many other word pairs (not in a capital-of relationship) whose difference is also similar to these differences \cite{ziedcoling}. Accordingly, it was found in \cite{Vylomova2016} that a relation classifier which is trained on word vector differences is prone to predicting many false positives. In contrast, we can expect that our relation vectors are modeling relations in a far less ambiguous way. On the other hand, these relation vectors are limited to word pairs that co-occur sufficiently frequently. Apart from the associated sparsity issues, this also suggests that relation vectors are not suitable for characterizing syntagmatic relationships and several types of syntactic relationships. We thus view these relation vectors as complementary to word vector differences.

%\subsubsection{Exploring Similar Relations}

%new for camera ready version

\begin{table}[!th]
\centering
\renewcommand{\arraystretch}{1.075}
%\resizebox{\textwidth}{!}{
\footnotesize
\begin{tabular}{cccc}

\multicolumn{4}{c}{\textbf{lime\_juice}}                                        
\\ \hline \hline
\texttt{original}             & \texttt{compressed-10d}          & \texttt{compressed-50d}        & \texttt{diffvec}               \\ \hline 
lemon\_juice         & lemon\_juice            & lemon\_juice          & lime\_soda            \\
juice\_lemon         & coconut\_milk           & juice\_lemon          & lime\_lemon           \\
juice\_lime          & marzipan\_paste         & juice\_lime           & lemon\_juice          \\
lime\_lemon          & juice\_lime             & vinegar\_sour         & citric\_juice         \\
lemon\_lime          & noodles\_egg            & lemon\_lime           & tamarind\_juice       \\
pineapple\_juice     & lime\_lemon             & vinegar\_sauce        & lime\_pie             \\
orange\_juice        & marinated\_beef         & lime\_lemon           & pineapple\_juice      \\\Xhline{3\arrayrulewidth}
\multicolumn{4}{c}{\textbf{nintendo\_console}}                                                          \\ \hline \hline
\texttt{original}             & \texttt{compressed-10d}          & \texttt{compressed-50d}        & \texttt{diffvec}               \\ \hline
wii\_console         & wii\_console            & wii\_console          & nintendo\_consoles    \\
playstation\_console & playstation\_console    & nintendo\_nes         & nintendo\_handheld    \\
nintendo\_nes        & nintendo\_nes           & playstation\_console  & gamecube\_console     \\
xbox\_console        & witcher\_2              & xbox\_console         & wii\_console          \\
nintendo\_consoles   & itunes\_download        & nintendo\_consoles    & dreamcast\_console    \\
famicom\_console     & imax\_2d                & sega\_consoles        & nintendo\_switch      \\
nintendo\_64         & netflix\_streaming      & nintendo\_handheld    & 3ds\_console          \\\Xhline{3\arrayrulewidth}
\multicolumn{4}{c}{\textbf{gmail\_email}}                                                               \\ \hline \hline
\texttt{original}             & \texttt{compressed-10d}          & \texttt{compressed-50d}        & \texttt{diffvec}               \\ \hline
yahoo\_email         & renders\_firefox        & yahoo\_email          & gmail\_emails         \\
inbox\_email         & ie\_browser             & gmail\_e-mail         & yahoo\_email          \\
hotmail\_email       & infinitive\_suffix      & inbox\_email          & hotmail\_email        \\
email\_yahoo         & firefox\_browser        & gmail\_emails         & addy\_email           \\
gmail\_e-mail        & carnap\_semantics       & email\_yahoo          & imap\_email           \\
sending\_email       & helvetica\_font         & hotmail\_email        & smtp\_email           \\
send\_email          & cv\_syllable            & google\_search        & bugzilla\_email       \\\Xhline{3\arrayrulewidth}
\multicolumn{4}{c}{\textbf{roman\_numerals}}                                                            \\ \hline \hline
\texttt{original}             & \texttt{compressed-10d}          & \texttt{compressed-50d}        & \texttt{diffvec}               \\ \hline
arabic\_numerals     & arabic\_alphabet        & arabic\_numerals      & cyrillic\_numerals    \\
letters\_numerals    & greek\_alphabet         & letters\_numerals     & indic\_numerals       \\
letters\_alphabet    & 10-inch\_discs          & uppercase\_letters    & georgian\_numerals    \\
lowercase\_letters   & latin\_alphabet         & lowercase\_letters    & hieratic\_numerals    \\
arabic\_alphabet     & yemenite\_pronunciation & uppercase\_characters & brahmi\_numerals      \\
latin\_alphabet      & standard\_orthography   & latin\_alphabet       & sinhala\_numerals     \\
symbols\_numerals    & wii\_remote             & alphabetic\_numerals  & quantifiers\_numerals \\\Xhline{3\arrayrulewidth}
\multicolumn{4}{c}{\textbf{heavy\_metal}}                                                               \\\hline \hline 
\texttt{original}             & \texttt{compressed-10d}          & \texttt{compressed-50d}        & \texttt{diffvec}               \\ \hline
thrash\_metal        & metal\_heavy            & thrash\_metal         & heavy\_metals         \\
glam\_metal          & karma\_dharma           & doom\_metal           & cky\_metal            \\
doom\_metal          & techno\_rave            & glam\_metal           & manilla\_metal        \\
symphonic\_metal     & psychedelic\_garage     & thrash\_slayer        & annihilator\_metal    \\
nu\_metal            & cooking\_recipes        & punk\_rock            & heaviness\_metal      \\
sludge\_metal        & gita\_yoga              & hardcore\_punk        & doro\_metal           \\
glam\_rock           & post-punk\_punk         & sludge\_metal         & behemoth\_metal      
\end{tabular}
%}
\caption{Nearest neighbors (by cosine) for selected relation vectors and the three models under consideration.}
\label{tab:neighbors}
\end{table}

In this section we illustrate the semantic properties of different versions of SeVeN. To this end, we show the nearest neighbors of selected target relation vectors for a number of different representations: (1) the original 1800d SeVeN network (\texttt{original}), (2) an autoencoded 10-dimensional space (\texttt{compressed-10d}), (3) a slightly higher-dimensional version (\texttt{compressed-50d}), and finally (4) a baseline model according to which the relation between two words is modeled as the vector difference of the corresponding word vectors
%consisting of subtracting and normalizing the individual word embeddings corresponding to each in the relation 
(\texttt{diffvec}). 
%of selected target relation vectors, in the two most extreme configurations of SeVeN, i.e.\ the original uncompressed 1800 dimensional and autoencoded 10 dimensional vectors. An overarching feature of these vectors is that relational properties of each vector become gradually more pronounced (and also more abstract and less domain-dependent) as the dimensionality shrinks, arguably because the autoencoder is more forced to learn to reconstruct the original relation vector without relying on the argument words that encode it. This phenomenon can be observed in that context and center words in lower dimensional spaces are more dissimilar than in the original one, while at the same time mostly preserving the gist of the relation.
The five selected target relation vectors, along with their nearest neighbors, are shown in Table \ref{tab:neighbors}.  These target relation vectors were chosen to capture a range of different types of relationships, including hypernymic (`nintendo - console' and `gmail - email') and attributional (`roman - numerals') relations. %,  highly attributive (almost collocational), e.g., in `roman-numerals', as well as relational (`lime-juice') or hypernymic (`nintendo-console' and `gmail-email'). For each of them, we show the 7 nearest neighbors in four dedicated vector spaces: 

One immediate observation is that, in most cases, the \texttt{diffvec} neighbors remain very close to the given word pair, where each word from the given pair is either preserved or replaced by a closely related word. The \texttt{original} and \texttt{compressed-50d} relation vectors largely follow a similar trend, although a few more interesting analogies are also found in these cases (e.g. \textit{arabic - alphabet} as a neighbor of \textit{roman - numerals}). The results for the \texttt{compressed-10d} vectors, however, follow a markedly different pattern. For these low-dimensional vectors, our autoencoder forces the relation vectors to focus on modeling the relationship between the two words, while abstracting away from the initial domain. This leads to several interesting neighbors, although this seems to come at the cost of some added noise.
%Let us focus on the properties that the \texttt{compressed-10d} space exhibits. We observe the trend of preserving (with arguably also a slight amount of unwanted noise) the gist of the relation encoded by the target vector, but with the added value of extrapolating the idea outside the initial domain, which we can verify because of the dissimilarity between each individual word vector in the relation with respect to their target counterparts. 

Let us now analyze more closely the results of the \texttt{compressed-10d} vectors. If we read the first example along the lines of ``juice can be made from limes'', similar relations are found close in the space, such as `coconut - milk' and `marzipan - paste'. Note that the relation `noodles - egg' is also similar, although the two words appear in the incorrect order (i.e.\ noodles can be made from eggs rather than the other way around). As another example where the directionality of this pattern is not captured correctly, we also find the pair `juice - lime'. It would be interesting to analyze in future work whether such issues can be avoided by using features from a dependency parser, e.g.\ following a similar strategy as in \cite{levy2014dependency}. Note that while all the \texttt{compressed-10d} neighbors are still related to food, these vectors have generalized beyond the domain of citrus fruits  (see e.g., `lime', `tamarind' or `lemon' in \texttt{diffvec}, or `lemon' and `orange' in \texttt{original}). A similar phenomenon occurs in some of the other examples. In the `nintendo-console' case, after interpreting the relation as ``major supplier of'' or ``entity which popularized'', we find nearest neighbors in the \texttt{compressed-10d} space where the same relation holds, but which do not belong to the video games domain, such as `itunes-download' or `netflix-streaming'. Next, we find that the relation holding between `gmail' and `email' is similar to `ie' and `firefox' and `browser', and even `helvetica' and `font'. The relation between `google' and `search', found for \texttt{compressed-50d}, is also of this kind. In contrast, the \texttt{diffvec} neighbors in this case all have \textit{email} as the second word. In the `roman-numerals' example, likewise, the \texttt{diffvec} neighbors similarly have `numerals' as the second word, while for \texttt{compressed-10d} we see more interesting neighbors such as `arabic-alphabet' and `yemenite-pronunciation'. We also find the seemingly unrelated `wii-remote' pair, although we may consider that the Nintendo Wii console introduced a fundamentally new type of remote, which at an abstract level is similar to the fact that the Romans introduced a fundamentally new way of writing numbers. This example also suggests, however, that the way in which relations are modeled in the 10-dimensional space might be too abstract for some applications. Finally, the `heavy-metal' case is a paramount example of how the the relation vectors may capture information which is fundamentally different than what is encoded by word vectors. In particular, the \texttt{diffvec} vectors all express relationships from the metalwork domain (e.g., `heavy-metals' or `annihilator-metal'), which reflects the fact that the music-related interpretation of the word `metal' is not its dominant sense. In contrast, since our relation vectors are exclusively learned from sentences where both words co-occur (`heavy' and `metal' in this example), the vector for `heavy metal' clearly captures the musical sense (see e.g., `thrash-metal' or `glam-metal' in the \texttt{original} space).

%\nb{Ideally, we'd really want a comparison here with vector differences, showing how noisy the nearest neighbors are if you only take these differences into account. While there is no time to do a full analysis of this, maybe you can still find an example of a word pair that is more similar to (rome,italy) than (paris, france). Zied recently found (for a different embedding) that the following pairs are in a "capital of" relation according to the word vector differences: (horses,horse), (girl,boy), (walks,walk), (unaware,aware), (looking,look). Another possible strategy for finding counterexamples is to flip the ordering, showing, say, that (france,paris) is more similar than (paris,france)}
%\nb{Again, it would be useful to compare these results with the original full-dimensional vectors, where we would hope to find that e.g. the most similar pair to (rome, italy) is perhaps (florence, italy). Again, of course there is only time to test a few examples, but maybe we can still find such a counterexample.}

% The high-low norm ranks didn't make too much sense to me. We'll have to inspect them better after this deadline.
%\todo{If there is still time/space, consider adding a table with examples of word pairs whose relation vector has a large L2 norm and pairs whose relation vector has a close-to-0 L2 norm.}

%--------------------
\subsection{Modeling Similarity}

%\url{http://groups.psych.northwestern.edu/gentner/papers/GoldstoneMedinGentner91.pdf} discusses the role of attributes vs relations in human similarity judgement (``This paper explores the alternative hypothesis that similarity models cannot adequately predict similarity judgments if relational features and simple features (attributes) are undifferentiated.'').

%not too sure about this par.
The capacity to capture and \textit{embed} nuances of word meaning is one of the most celebrated features of word embeddings. The task of semantic similarity measurement, therefore, has been adopted as a \textit{de-facto} testbed for measuring the quality of representations of linguistic items. The standard practice is to consider a distance (or similarity) metric such as cosine similarity and compare the similarity in a given vector space model with respect to human judgement. We note, however, that there exist other similarity metrics discussed in the literature, e.g., Weighted Overlap \cite{pilehvar2013align} or Tanimoto Distance \cite{iacobacci2015sensembed}. Our proposed similarity measurement parts ways from the idea of improving the representation of individual words, and rather seeks to refine their meaning by incorporating complementary cues via relation vectors, as well as the corresponding neighborhood structure. There are many possible ways in which this could be done, but we restrict ourselves here to a simple strategy, based on identifying the closest neighbors of the two words $w_1$ and $w_2$. The main intuition is that when $w_1$ and $w_2$ are similar, they should also be related to similar words.
Specifically, we first determine the closest match between the neighbors of $w_1$ and the neighbors of $w_2$, as follows:
\begin{align*}
(n_1,n_2) = \argmax_{(n_1,n_2) \in N_{w_1} \times N_{w_2}} \cos(\mathbf{v}_{n_1},\mathbf{v}_{n_2})+\cos(\mathbf{r}_{w_1n_1},\mathbf{r}_{w_2n_2})
\end{align*}
Note that to identify these neighbors, we compare both their word vectors $\mathbf{v}_{n_1}$ and $\mathbf{v}_{n_2}$, and their relationships to the target words, $\mathbf{r}_{w_1n_1}$ and $\mathbf{r}_{w_2n_2}$. Once these neighbors have been identified, we compute the similarity between $w_1$ and $w_2$ as follows:
\begin{align*}
\textit{sim}(w_1,w_2) = \cos(\mathbf{v}_{w_1}\oplus \mu \mathbf{v}_{n_1} \oplus \mathbf{r}_{w_1n_1},\mathbf{v}_{w_2}\oplus \mu\mathbf{v}_{n_2}\oplus \mathbf{r}_{w_2n_2}) 
\end{align*}
where $0<\mu\leq 1$ is a scaling factor which is aimed at reducing the impact of the neighbors $n_1$ and $n_2$ on the overall similarity computation. The fact that $\mathbf{v}_{n_1}$ is similar to $\mathbf{v}_{n_2}$ is an important indicator for the similarity between $w_1$ and $w_2$, but it should not influence the resulting similarity score as much as the similarity of the word vectors of $w_1$ and $w_2$ themselves. Rather than tuning this value, in the experiments we have fixed it as $\mu=0.5$, which was found to give better results than $\mu=1$ (i.e.\ no scaling). Note that the proposed way of computing similarities favours words of the same type. For example, we may expect `Spain' and `France' to be more similar than `Spain' and `Barcelona', when this metric is used, since `Spain' and `France' are associated with the neighbors `Madrid' and `Paris' which are similar, and which are related in a similar way to the target words. In our experiments, we also consider a variant in which the relation vectors are only used for selecting the neighbors. The similarity itself is then calculated as:
\begin{align*}
\textit{sim}(w_1,w_2) = \cos(\mathbf{v}_{w_1}\oplus \mu \mathbf{v}_{n_1},\mathbf{v}_{w_2}\oplus \mu\mathbf{v}_{n_2}) 
\end{align*}

\noindent We evaluate the proposed similarity measure on four well-known benchmarking datasets for word representation learning. These are: (1) \texttt{rg65} \cite{rubenstein1965contextual}; (2) \texttt{wordsim} \cite{finkelstein2001placing}; (3) \texttt{mc} \cite{miller1991contextual}; and (4) the English portion of \texttt{semeval17} \cite{camacho2017semeval}. We restrict our experiment to single words, and do not consider multiword expressions (e.g., named entities), as this would require a different approach for compositional meaning representation. We compare against a baseline model based on cosine similarity between the vectors of the target words (\texttt{cosine}). As for our proposed models, and observing the similarity measurement described above, we consider a 10-dimensional relation space, without (10rv$_{w}$) and with (10rv$_{r}$) the relation vector as part of the similarity computation. We also provide results stemming from using the original 1800-dimensional relation vector model. As is customary in the literature, we use Pearson's (\textbf{p}) and Spearman's (\textbf{s}) correlation coefficients as evaluation metrics, as well as their average (\textbf{avg.}). Table \ref{tab:wordsim} shows that the 10rv$_{w}$ variant consistently outperforms the word-level baseline. Somewhat surprisingly, the variant 10rv$_{r}$ (which uses the relation vector also in the similarity computation) performs consistently worse than the variant 10rv$_{w}$. When using the original 1800-dimensional vectors, however, the situation is reversed, with 1800rv$_{r}$ outperforming 1800rv$_{w}$, and achieving the best results overall (with the exception of \texttt{mc}). These results clearly show that the relation vectors capture valuable information for measuring word similarity, although the information captured by the 10-dimensional vectors may in some cases be too abstract for this purpose.

%While the improvements are overall relatively small, they clearly show that incorporating relational information improves similarity judgements. A potential drawback of incorporating relational information in the similarity computation may be that it may ``pull apart'' similar words, which however may have not a strong relational bond. In conclusion, however, while there is certain variability in performance when considering 10 or 1800-dimensional vectors, and when factoring in the relation vector for similarity computations, it seems clear that the SeVeN vectors are strongly contributing to overall better performance across all datasets and all considered metrics.

%This could explain the slightly worse performance of the ``relational'' equation in e.g., \texttt{rg} and \texttt{mc} (which are also the smallest datasets, and also more prone to be affected by outliers). 

%Further gains can be expected by e.g.\ tuning a weighted balance between the vectors $\mathbf{v}_{w_i}$ and $\mathbf{r}_{w_in_i}$ in the similarity computation, and by looking at the top-k closest matches, rather than the single closest match, when determining which edge(s) to taken into account.
%Note that the performance of existing state-of-the-art systems is as of today at par with or even better in some cases than human inter-annotator agreement \cite{camacho2017semeval}.

\begin{table}[!h]
\centering
\small
\renewcommand{\arraystretch}{1.2}
\resizebox{\textwidth}{!}{%
\begin{tabular}{lrrr|rrr|rrr|rrr} 
                                 & \multicolumn{3}{c|}{\texttt{rg}}                                                    & \multicolumn{3}{c|}{\texttt{wordsim}}                                              & \multicolumn{3}{c|}{\texttt{mc}}                                                   & \multicolumn{3}{c}{\texttt{semeval17}}                                           \\ \cline{2-13} 
                                 & \multicolumn{1}{c}{\textbf{p}} & \multicolumn{1}{c}{\textbf{s}} & \multicolumn{1}{c|}{\textbf{avg.}} & \multicolumn{1}{c}{\textbf{p}} & \multicolumn{1}{c}{\textbf{s}} & \multicolumn{1}{c|}{\textbf{avg.}} & \multicolumn{1}{c}{\textbf{p}} & \multicolumn{1}{c}{\textbf{s}} & \multicolumn{1}{c|}{\textbf{avg.}} & \multicolumn{1}{c}{\textbf{p}} & \multicolumn{1}{c}{\textbf{s}} & \multicolumn{1}{c}{\textbf{avg.}} \\ \hline \hline
\multicolumn{1}{l|}{\texttt{cosine}}       & 77.2                  & 76.0                  & 76.6                      & 64.9                  & 69.4                  & 67.1                      & 79.2                  & 80.0                  & 79.6                      & 69.4                  & 70.0                  & 69.7                     \\ \hline
\multicolumn{1}{l|}{10rv$_{w}$} & 78.1 & 77.0 & 77.5 & 66.0 & 69.6 & 67.8 & 79.7 & 80.7 & \textbf{80.2} & 70.2 & 70.8 & 70.5 \\
\multicolumn{1}{l|}{10rv$_{r}$} & 77.4 & 75.5 & 76.4 & 65.8 & 69.5 & 67.6 & 78.8 & 77.9 & 78.3 & 70.0 & 70.7 & 70.3                     \\
\multicolumn{1}{l|}{1800rv$_{w}$}     & 79.5                  & 80.6                  & \textbf{80.0}                      & 67.4                  & 69.8                  & 68.6                      & 79.4                  & 79.0                  & 79.2                      & 71.4                  & 71.8                  & 71.6                     \\
\multicolumn{1}{l|}{1800rv$_{r}$}     & 78.9 & 80.2 & 79.5 & 68.1 & 70.1 & \textbf{69.1} & 79.2 & 79.7 & 79.4 & 72.2 & 73.0 & \textbf{72.6}                 
\end{tabular}
}
\caption{Correlation results for different configurations of our proposed approach and a competitor baseline based on cosine similarity of word embeddings.}
\label{tab:wordsim}
\end{table}

%&To illustrate how the relation vectors help to improve similarity judgments, let us focus on the word pair ... (add new example)

%(\textit{penny}, \textit{bargain}), which is included in the \texttt{semeval17} dataset with a gold score of 1.42 (out of 6). Our proposed metric can be seen as a way to query the relation space for words that co-occur often with these two and are at the same time informative. In our relation graph, we find \textit{make} as the highest scoring relation between both words (i.e., the relation vectors $\mathbf{r}_{\textit{make}\,\textit{penny}}$ and $\mathbf{r}_{\textit{make}\,\textit{bargain}}$), and consider their semantic similarity to act as an increasing or decreasing factor towards the final decision. In this case, the initial word similarity of 0.39 is decreased to 0.37, reflecting the fact that there is no clear relational similarity between the pairs (make,penny) and (make,bargain). The opposite happens for the pair (\textit{church}, \textit{cathedral}), where the closest match links both words to the neighbor \textit{st}\footnote{Meaning `saint'.}. Here, the relational information acts as a similarity booster (from 0.71 to 0.75) because the vectors $\mathbf{r}_{\textit{church}\,\textit{st}}$ and $\mathbf{r}_{\textit{cathedral}\,\textit{st}}$ are rather similar.

\subsection{Text Classification}

Semantic Vector Networks may be thought of as a natural way of enriching word-level semantic representations, which may in turn be useful for informing a neural architecture with relational (e.g., commonsense or lexical) knowledge. We will focus on two well known tasks, namely text categorization and sentiment analysis. Our goal is to examine the extent to which the performance of a vanilla neural network increases by being injected vector graph information as a complement to the information encoded in each individual word embedding. The strength of our proposal lies in the fact that this information comes exclusively from corpora, and thus the need to rely on often incomplete, costly and language dependent ontological or lexical resources is avoided.

%\begin{center}
%\begin{figure}[!h]
%\centering
%\includegraphics[width=10cm,height=5.5cm]{input-2-small}
%\caption{The input layer leveraging semantic vector networks. It consists of a max pooling layer on top of a matrix encoding neighborhood and relational information for each sentence token.}
%\label{fig:model}
%\end{figure}
%\end{center}

%In Figure \ref{fig:model} we illustrate how our vector networks can be fed to a neural network, where we simply replace each word embedding by the word's associated subgraph. 

As evaluation benchmarks we use 
%the \texttt{bbc} \cite{greene2006practical} topic categorization corpus,
three text categorization datasets, namely \texttt{20news} \cite{lang1995newsweeder}, \texttt{bbc} \cite{greene2006practical} and \texttt{reuters} \cite{lewis2004rcv1}. We also consider two polarity detection datasets (positive or negative), namely the Polarity04 (\texttt{pol.04}) \cite{pang2004sentimental} and Polarity05 (\texttt{pol.05}) \cite{pang2005seeing} datasets, and finally a 10k document subset of the \textit{apps for android} (\texttt{apps4and.}) corpus\footnote{Obtained from \url{http://jmcauley.ucsd.edu/data/amazon/}.} \cite{he2016ups}, which features reviews and associated ratings on a scale from 1 to 5. The neural network model we use for our experiments is a combination of a CNN \cite{LeCunnetal1998} and a bidirectional LSTM \cite{HochreiterandSchmidhuber1997}. CNNs have been evaluated extensively in text classification \cite{johnson2014effective,tang2015document,xiao2016efficient,conneau2017very} and sentiment analysis \cite{kalchbrenner2014convolutional,Kim2014,dos2014deep,yin2017comparative}, and this specific model (CNN+BLSTM) has been explored in different NLP benchmarks \cite{Kim2014}. Finally, as evaluation metrics we use precision (\textbf{p}), recall (\textbf{r}) and f-score (\textbf{f}), as well as accuracty (\textbf{acc.}).

To use SeVeN for text classification, we keep the exact same neural network architecture, but use enriched vector representations for each word. As a proof-of-principle, in this paper, this enriched vector representation is simply obtained by concatenating the word vector of that word, with vector representations of its top-10 neighbors according to PMI (ordered by this PMI score), together with the corresponding relation vectors.
%The graph representation that replaces the embedding of a word is simply given by a vector which concatenates to the word embedding of such word a sequence of relation vectors and corresponding neighboring word, to a maximum of 10, sorted by PMI score. 

For example, with word embeddings of 300 dimensions and relation vectors of 10 dimensions, the input for each word is given by a 3,400-dimensional vector. We list experimental results for several configurations, where the number of neighbors stays fixed, but the relation vector (rv) changes in dimensionality (10, 20 or 50). Experimental results are provided in Table \ref{tab:classification}. We can see that for the 20-dimensional vectors, the results are consistently better, or at least as good as the \texttt{baseline}. The results for the 10-dimensional and 50-dimensional vectors are similar, although these configurations perform slightly worse than the baseline for \texttt{pol.05}.

Overall, these results show the usefulness of the relation vectors and neighborhood structure, despite the rather naive way in which this information is used. It seems plausible to assume that performance may be further improved by using network architectures which exploit the graph structure in a more direct way.

\begin{table}[!t]
\centering
\resizebox{\textwidth}{!}{
\renewcommand{\arraystretch}{1.15}
\begin{tabular}{lrrr|rrr|rrr|rrr|r|r}
           & \multicolumn{3}{c|}{\texttt{bbc}}                                               & \multicolumn{3}{c|}{\texttt{20news}}                                             & \multicolumn{3}{c|}{\texttt{reuters-r56}}                                       & \multicolumn{3}{c|}{\texttt{apps4and.}}                                  & \multicolumn{1}{c|}{\texttt{pol.04}} & \multicolumn{1}{c}{\texttt{pol.05}} \\ \cline{2-15} 
           & \multicolumn{1}{c}{\textbf{p}} & \multicolumn{1}{c}{\textbf{r}} & \multicolumn{1}{c|}{\textbf{f}} & \multicolumn{1}{c}{\textbf{p}} & \multicolumn{1}{c}{\textbf{r}} & \multicolumn{1}{c|}{\textbf{f}} & \multicolumn{1}{c}{\textbf{p}} & \multicolumn{1}{c}{\textbf{r}} & \multicolumn{1}{c|}{\textbf{f}} & \multicolumn{1}{c}{\textbf{p}} & \multicolumn{1}{c}{\textbf{r}} & \multicolumn{1}{c|}{\textbf{f}} & \multicolumn{1}{c|}{\textbf{acc.}}       & \multicolumn{1}{c}{\textbf{acc.}}       \\ \hline \hline
\texttt{baseline}     & 0.95                  & 0.95                  & \textbf{0.95}                   & 0.86                  & 0.85                  & 0.86                   & 0.85                  & 0.88                  & 0.86                   & 0.39                  & 0.48                  & 0.38                   & 0.54                            & \textbf{0.78}                           \\ \hline
10rv & 0.95                  & 0.95                  & \textbf{0.95}                   & 0.88                  & 0.87                  & 0.87                   & 0.89                  & 0.91                  & \textbf{0.90}                   & 0.40                     & 0.44                     & \textbf{0.40}                      & 0.56                            & 0.75                           \\
20rv & 0.96                  & 0.95                  & \textbf{0.95}                   & 0.89                  & 0.89                  & \textbf{0.89}                   & 0.89                  & 0.92                  & \textbf{0.90}                   & 0.38                     & 0.48                     & \textbf{0.40}                      & 0.59                            & \textbf{0.78}                           \\
50rv & 0.94                  & 0.94                  & 0.94                   & 0.88                  & 0.87                  & 0.88                   & 0.89                  & 0.91                  & \textbf{0.90}                   & 0.35                     & 0.46                     & 0.38                      & \textbf{0.60}                            & 0.77                          
\end{tabular}
}
\caption{Experimental results on six benchmarking datasets for text classification.}
\label{tab:classification}
\end{table}

%\begin{table}[!h]
%\centering
%\begin{tabular}{lcrr|crr|crr}
%         & \multicolumn{3}{c|}{\textbf{bbc}}                                         & \multicolumn{3}{c|}{\textbf{polarity04}}                                  & \multicolumn{3}{c}{\textbf{polarity05}}                                  \\ \cline{2-10} 
%         & P                        & \multicolumn{1}{c}{R} & \multicolumn{1}{c|}{F} & P                        & \multicolumn{1}{c}{R} & \multicolumn{1}{c|}{F} & P                        & \multicolumn{1}{c}{R} & \multicolumn{1}{c}{F} \\ \hline
%Baseline & \multicolumn{1}{r}{0.72} & 0.71                  & 0.71                   & \multicolumn{1}{r}{0.54} & 0.53                  & 0.54                   & \multicolumn{1}{r}{0.55} & 0.54                  & 0.54                  \\ \hline
%SeVeN    & \multicolumn{1}{r}{\textbf{0.78}} & \textbf{0.77}                  & \textbf{0.77}                   & \multicolumn{1}{r}{\textbf{0.81}} & \textbf{0.54}                  & \textbf{0.61}                   & \multicolumn{1}{r}{\textbf{0.66}} & \textbf{0.63}                  & \textbf{0.64}                 
%\end{tabular}
%\caption{Experimental results on benchmarking datasets for text classification. Baseline refers to a CNN+BLSTM model with a first embedding layer. SeVeN is the same model, but taking as input the matrix of each word's neighborhood.}
%\label{tab:classification}
%\end{table}

%******************************************************************************************************************************
\section{Conclusions and Future Work}

In this paper we have presented SeVeN, a dedicated vector space model for relational knowledge. These relation vectors encode corpus-based evidence capturing the different contexts in which a pair of words may occur. An initially high-dimensional relation vector is is further ``purified'' thanks to a simple \textit{ad-hoc} autoencoder architecture, designed to only retain relational knowledge. We have explored the characteristics of these vector networks qualitatively, by showing highly correlated word pairs, as opposed to, for example, difference vectors. While the latter are often assumed to capture relational properties, we found that the relational similarities they capture largely reflect the similarities of the individual words, with little relational generalization capability. In addition, we have evaluated our SeVeN vectors in terms of their usefulness in two standard NLP tasks: word similarity and text classification. In both cases we obtained better results than baselines that use standard word vectors alone.

There are several interesting avenues for future work. First, an obvious way to improve these unsupervised representations would be to leverage structured knowledge retrieved from knowledge graphs and/or Open Information Extraction systems. Such knowledge could easily be exploited by feeding any available structured knowledge as additional inputs to the autoencoder. Another way in which structured knowledge could be harnessed would simply be to label relation vectors, i.e.\ to identify regions in the relation vector space that correspond to particular relation types (e.g.\ hypernymy). Another possibility would be to improve SeVeN by aggregating relation vectors along paths in the graph. In this way, we may learn to predict missing edges (or to smooth relation vectors that were learned from too few or too uninformative sentences), similarly to the random walk based strategies that have been developed for completing traditional semantic networks and knowledge graphs \cite{DBLP:conf/emnlp/GardnerTKM14}.

\section*{Acknowledgments}
We would like to thank the anonymous reviewers for their helpful comments. This work was supported by ERC Starting Grant 637277.

%Future work:
%\begin{itemize}
%\item Learn to predict missing edges (e.g.\ by aggregating path between two words using recurrent net)
%\item Use input from knowledge graphs
%\end{itemize}

\bibliographystyle{acl}
\bibliography{commonsense,wordembedding}

\end{document}